\pdfoutput=1
\documentclass[letterpaper, 10 pt, conference]{ieeeconf}  % Comment this line out if you need a4paper

\IEEEoverridecommandlockouts                              % This command is only needed if 
\usepackage{graphicx} % for pdf, bitmapped graphics files
\usepackage{epstopdf} % for postscript graphics files
\usepackage{dcolumn}

\title{\LARGE \bf
Towards an objective evaluation of underactuated gripper designs
}

\author{Eduardo Ruiz and Walterio Mayol-Cuevas% <-this % stops a space
\thanks{The authors are with the Department of Computer Science,
        University of Bristol, UK
        {\tt\small er13827@bristol.ac.uk, wmayol@cs.bris.ac.uk}}%
}

\begin{document}

\maketitle
\thispagestyle{empty}
\pagestyle{empty}

%involving tools that cooperate with the human in the accomplishment of multiple tasks, a concept that had been relatively unexplored until recently, handheld robots. 

%%%%%%%%%%%%%%%%%%%%%%%%%%%%%%%%%%%%%%%%%%%%%%%%%%%%%%%%%%%%%%%%%%%%%%%%%%%%%%%%
\begin{abstract}

In this paper we explore state-of-the-art underactuated, compliant
robot gripper designs through looking at their performance on a
generic grasping task. Starting from a state of the art open gripper
design, we propose design modifications, and importantly, evaluate all
designs on a grasping experiment involving a selection of objects
resulting in 3600 object-gripper interactions.  Interested in
non-planned grasping but rather on a design's generic performance, we
explore the influence of object shape, pose and orientation relative
to the gripper and its finger number and configuration. Using
open-loop grasps we achieved up to 75\% success rate over our
trials. The results indicate and support that under motion constraints
and uncertainties and without involving grasp planning, a
2-fingered underactuated compliant hand outperforms higher
multi-fingered configurations. To our knowledge this is the first
extended objective comparison of various multi-fingered underactuated
hand designs under generic grasping conditions.
\end{abstract}

%%%%%%%%%%%%%%%%%%%%%%%%%%%%%%%%%%%%%%%%%%%%%%%%%%%%%%%%%%%%%%%%%%%%%%%%%%%%%%%%
\section{INTRODUCTION}

% \begin{itemize}
% \item Grab most objects
% \item Performance/Success based measure
% \item Under-actuated compliant hand
% \item Simulation-only compared to real-scene experiment
% \item Different object classes not necessarily ideal
% \end{itemize}

Robotic platforms requiring to interact with objects have made use of
a wide range of end-effectors. From parallel-jaw grippers and
simplified multifingered hands to highly articulated humanoid hands
remain among the most widely used.
   
In spite of numerous advances in the development of {\it general
  purpose} robotic hands, most of the designs are based on designers'
assumptions and pre-conceptions which are often hard to verify when
the system is implemented in a real robot; among other problems this
translates in developments struggling to receive wider acceptance
among researchers. In the specific case of robot hands, these are
often designed by starting with a high-level organizing principle that
defines the hand’s main features. An example of this are the
anthropomorphic hands, which are based on the idea that by reproducing
a human hand the robot will be able to perform human tasks. Another
example are the designs which build-up upon strong mathematical
principles using parameters such as force, workspace or contact area.
\begin{figure}[thpb]
      \centering
      \includegraphics[width=0.4\textwidth]{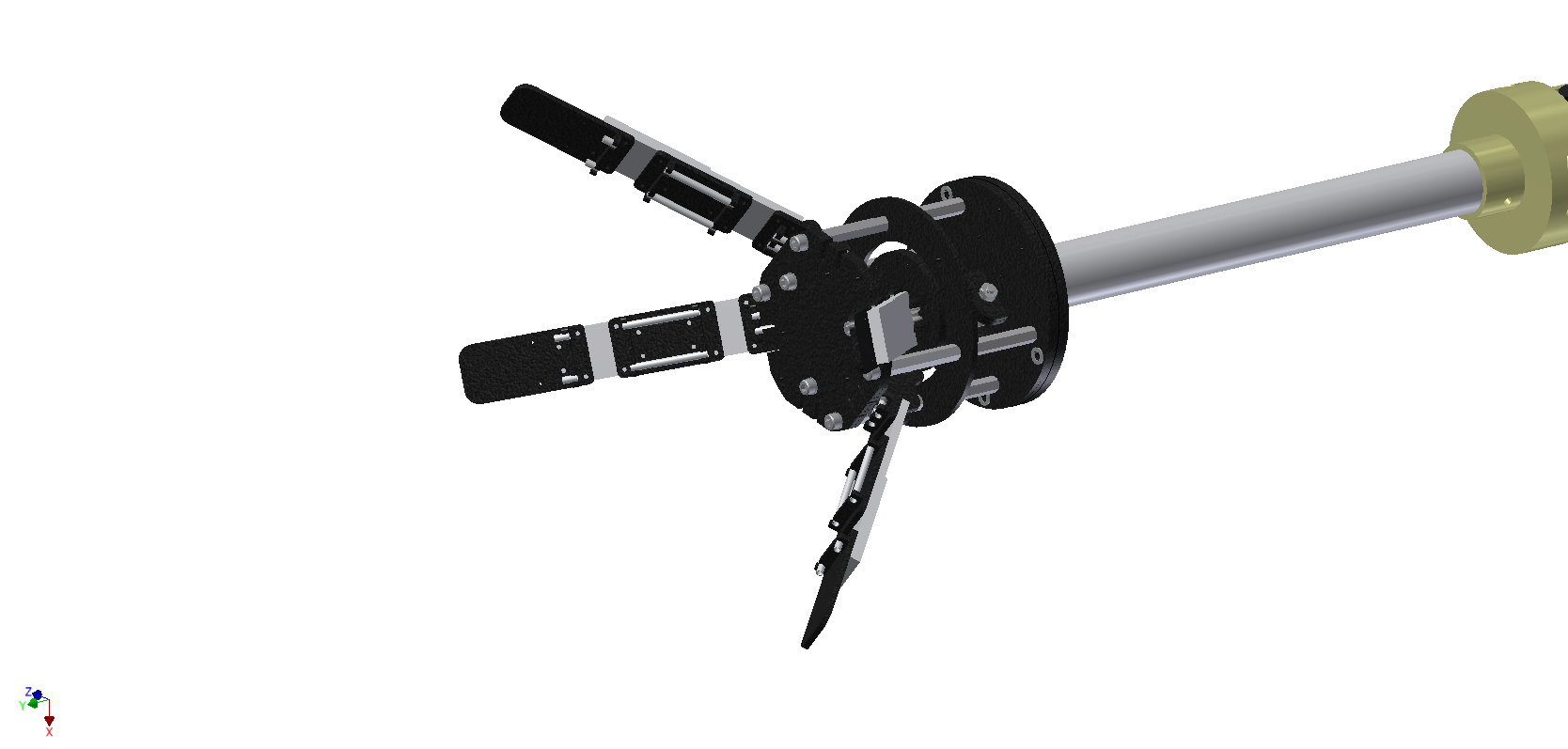}
      \caption{Variation of the hand configurations (4-fingers) used for the grasping experiments.}
      \label{fig: hand CAD}
\end{figure}

Although some systems have achieved impressive results in specific
circumstances, there is always the issue of balancing design
complexity, generality and or performance. Intuitively, an increased
number of fingers offers a higher number of contact points which can
arguably lead to a more robust grasp; however, it usually comes with
an increased control complexity and high manufacturing cost. On the
other hand, reducing the number of fingers or degrees of freedom
brings lower complexity and relatively lower cost, with the downside
of possibly reducing the dexterity of the system or limiting the
objects that the hand can grasp.

Until recent years, robotic grasping was dominated by analytic
approaches that took into account simplified 3D object models, complex
dynamics, contact-point information, etc. that allowed to compute high
quality or optimal grasp configurations for diverse objects
\cite{roa2014}. On the other hand, many recent approaches have used a
different approach where the grasp synthesis rely on sampling candidates
for and object, ranking them according to a specific metric using
grasp experience generated through demonstrations \iffalse
\cite{detry2013,lin2014,goins2014} \fi, trial and error \iffalse
\cite{kraft2010,barck2009,schiebener2012}\fi, labeled data \iffalse
\cite{zhao2014,goins2014,lenz2015}\fi and heuristics
\cite{bohg2014}. \iffalse\cite{fischinger2013}\fi
   
As can be seen in
\cite{roa2014,goins2014,laaksonen2010,bekiroglu2011}, most
state-of-the-art robotic grasping approaches make use of quality
metrics that require information such as hand joint positions, applied
forces/torques and contact points. On the contrary, real robotic
systems often face challenging situations which include uncertainties
in several aspects of the grasping task. In this type of scenario it
is often not possible to compute a reliable grasp metric; instead, a
metric based on the performance of a the grasp results more
suitable. While the most recent commercially available robotic hands
have plenty of sensors to provide all sorts of information, these
hands result expensive and require complex control algorithms.

%; in the same way, to design and build one of such devices is time
%consuming and at high cost.

In response to these challenging issues, underactuated hands offer the
advantage of requiring less actuators to control degrees of freedom,
having a lower complexity and in most cases a low manufacturing
cost. Moreover, adding compliance to an underactuated hand offers
advantages for working under uncertainty as joint positions do not
need to be exactly known. Research has focused on the use of design
parameters like: limits of the exerted forces, configuration of the
workspace, degrees of freedom and anthropomorphism. Whereas these have
proved to work to certain degree, one can also note that most usually
they are tested using reduced object sets or larger sets containing
multiple objects of the same classes. Designs are usually put to test
in a proof-of-concept setup rather than an extended scenario showing
the variety and amount of objects that the device can actually grasp.

In this paper, we carried out experiments to investigate the grasping
behaviour and performance of an underactuated compliant hand using
different finger configurations in a setup comprising of shelves in 3
levels. The experiment allowed to evaluate and compare the performance
of the grasps using objects with variable geometries and sizes, and
how this performance is affected when the number and configuration of
fingers is changed. All of this with an extensive number of grasp
trials.  To the best of our knowledge there has not been a study that
investigates underactuated grasping behaviour in this fashion.

The paper is organized as follows: Section \ref{section: context}
describes the compliant approach to robotic grasp. Section
\ref{section: model} describes the hand design used for the
experiments. Section \ref{section: experiment} presents the robotic
setup and experimental results of our study. Finally, Section
\ref{section: results} summarizes our findings and Section
\ref{section: conclusions} our conclusions.

\section{BACKGROUND}
\label{section: context}

% \begin{itemize}
%  \item Underactuated hands
%  \item Compliant grasping
%  \item Metrics in underactuated compliant grasping
%  \item Number of fingers
% \end{itemize}
\subsection{Underactuation and Compliance in robotic grasping}

Underactuated compliant hands leverage passive mechanisms and joint
coupling to reduce the number of actuators required to achieve robust
grasps. This reduction in the actuation comes with the benefits of
decreasing cost and improving planning efficiency.  As stated before,
there is always the trade-off among cost, complexity and effectiveness
when it comes to choosing the end effector in robotic grasping
applications. Designing this type of hand typically involves some
initial choices involving numbers and arrangements of fingers, types
of actuators, etc. followed by parametric variations of finger
lengths, transitions ratios, joint angles, often seeking to optimize
the hand using a standard grasping metric
\cite{dollar2010,hammond2012, aukes2012,aukes2014}.

Recent efforts attempting to join underactuation with compliance in
the hand mechanics have allowed to further simply topologies of
robotic hand actuation, at the same time affording passive adaptation
to unstructured settings and providing robustness against sensing
uncertainties \cite{dollar2010,odhner2011, malvezzi2013,
  odhner2014}. In this way, robot hands have increased robustness to
unexpected collisions with obstacles or unplanned contacts while
grasping objects.

Compliant underactuated hands have thus prompted a wide variety of
designs e.g.
\cite{barret2000,odhner2014,zisimatos2014,odhner2011,dollar2010}, most
of which show remarkable results in being able to grab unknown objects
robustly.

Dollar and Howe \cite{dollar2010}, presented the SDM Hand, a
4-fingered hand capable of grasping objects of regular geometries such
as prism, cylinders and spheres. In \cite{odhner2014, aukes2014}, 3
and 4-fingered hands are introduced, both of them with complex
actuation and sensory systems designed to met the DARPA ARM-H
challenge requirements; therefore, they are intended to perform well
defined tasks and grab specific classes of objects. \cite{odhner2012}
introduces a hand design for precision grasp and manipulation of small
objects placed in a table. Inspired by \cite{dollar2010}, Zisimatos et
al. in \cite{zisimatos2014} developed a ``general purpose'' robotic
hand with some important advantages such as modularity, low cost and
low weight; however, only preliminary results are reported.

We chose to follow the open design proposed in \cite{zisimatos2014},
which represents a good trade-off between cost, complexity and ability
to grab objects. A few modifications are introduced in order to adapt
such prototype to our requirements and to further test the grasp
reliability, these are detailed in Section \ref{section:
  experiment}. The hand design used for this study is shown in
Fig. \ref{fig: hand CAD}. Briefly speaking, this design has the
following characteristics: i)compliant flexure joints, ii)modular
basis with multiple slots, iii) differential disk mechanism
(underactuation), iv) low cost parts and v) cable driven actuation.

\subsection{Metrics and Performance}

When assessing the grasp performance or quality of a robotic system,
evaluation metrics usually fall into two broad categories: location of
the contact points on the object and hand configuration. These metrics
take into account properties such as disturbance resistance,
dexterity, equilibrium and stability. In essence, given an object and
a hand there are many possible grasps that satisfy a desired goal, an
optimal grasp is chosen using a quality metric that measures the
\textit{goodness} of that grasp. However, most of the widely accepted
metrics assume fully actuated multifinger hands, very few of them
have been adapted or developed for underactuated hands
\cite{roa2014}. And methods intended to optimize underactuated
grasping \cite{hammond2012,aukes2014}, are based on metrics that
assume a well defined contact model and optimized through simulations.
When the hands are built, designs are tested against prototypical
objects such as cylinders or prisms \cite{dollar2010, odhner2011,
  odhner2014} or by measuring the pulling force that the hand is able
to resist \cite{odhner2014, aukes2014,aukes2012}. In general, robot
grasp planning and automatic generation of grasps has been done using
performance metrics not stable when applied to real environments; this
makes it difficult to reproduce the planned grasp configuration with a
real robot hand, making the quality evaluation less useful in reality.

\begin{figure*}[thpb]
      \centering
      \includegraphics[width=\textwidth]{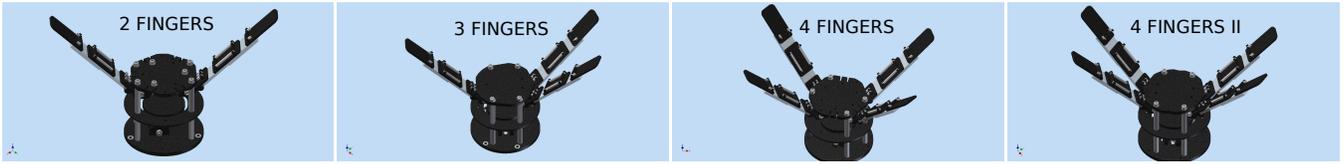}
      \caption{Hand configurations used for this experiment. From left to right: 2-finger, 3-fingers, 4-fingers and 4-fingers II.}
      \label{fig: hands cad}
\end{figure*}

When considering compliance into the grasp performance measure the
outlook remains somewhat the same. Researchers have adopted simplified
2D models \cite{malvezzi2013,grioli2012,aukes2012,odhner2011},
simulations \cite{borras2013, kim2012} or a combination of them. Other
works have used mathematical models that take into account stiffness
matrices, rigid body models, friction coefficients and contact
models. All of these studies apply, to some extent, standard grasp
metrics in order to evaluate algorithms and hand designs.

Due to the fact that most hand designs make use of kinematic models
and contact-based or wrench-space metrics, it has been showed that a
higher number of fingers provide more stable and reliable grasps. This
can be seen on earlier studies \cite{liu2000}, where the necessary
condition to stable grasp are deducted. This kind of analysis is based
in concepts as form closure and force closure, where the grasp hold by
the multifingered hand can cancel external wrenches applied to the
object, either using finger positions (form-closure) or forces
(force-closure). In spite of this, state-of-art robotic grasping
approaches have made use of a wide spectrum of devices, from 2-finger
grippers up to highly articulated 5-fingered hands.

We focus on studying the performance of grasps achieved in a real,
non-simulation-based setting, involving uncertainties in object pose
and modifying the number of fingers (contact points) involved in the
grasp. We believe that before moving to grasping algorithms that add
further intelligence when trying to grab objects, we must make sure of
using a device that actually can grab most objects. For that matter,
we investigate the performance of the state-of-the-art robotic hand
described in Section \ref{section: model}; using a setup of what we
believe approaches a commonly occurring real robotic scenario: a
multi-level shelf object picking task. Furthermore, since we want to
measure to what extent the hand is able to grasp objects ``by
itself'', we have reduced the degrees of freedom for hand/wrist
placement and conducted open-loop grasp experiments. In this way, we
could evaluate the grasping performance prior to any intelligence or
grasp planning algorithm.
 
\section{HAND DESIGN}
\label{section: model}

The hand design implemented for this study is based on the model
introduced by \cite{zisimatos2014}. A few modifications have been made
in order to adapt this hand to our framework. As first modification,
the actuator was removed from the end-effector module. Instead, the
motor is translated to the centre of mass of the entire tool, adding a
pulley system and using a cable driven actuation principle. In this
way we can balance the prototype and reduce the weight/load located at
the end effector. Having a light-weight end-effector potentially
allows us to work with heavier objects reducing torques on the rest of
the structure. The design and construction shares elements from
\cite{zisimatos2014}. The hand palm plate, differential disk and
finger phalanges are 3D printed. The finger joints are made of
silicone and the system is assembled together using plastic fasteners
to further decrease the hand weight.

\begin{figure}[thpb]
      \centering
      \includegraphics[width=0.4\textwidth]{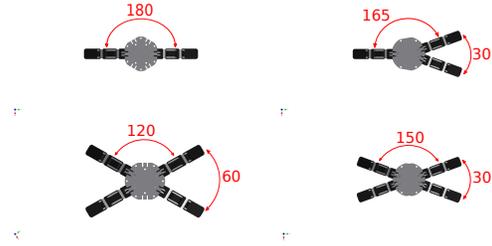}
      \caption{Hand configurations with annotated values for angle between pairs of fingers.}
      \label{fig: fingers}
\end{figure}

A further addition to the design is the inclusion of a new palm plate
set, this includes 2 new finger configurations: a cylindrical
configuration with two fingers opposing a third one, and a cylindrical
configuration with four fingers (two opposite pairs) forming a rather
spherical grasp. The finger configurations tested in this study are
shown in Fig. \ref{fig: hands cad}. To clarify the difference among
the configuration of the fingers Fig. \ref{fig: fingers} depicts the
hands in top-view, with annotated angles between the fingers.

Using the aforementioned design we explore the diversity of objects
and poses in which different finger configurations are able to obtain
a successful grasp. Furthermore, with the experimental procedure
employed we are able to gain insight about which hand configuration
performs best according to an object's geometry and pose. In the next
section we describe our experiment setup.

\section{EXPERIMENTAL PROCEDURE}
\label{section: experiment}
%\subsection{System overview}

\begin{figure}[thpb]
      \centering
      \includegraphics[width=0.4\textwidth]{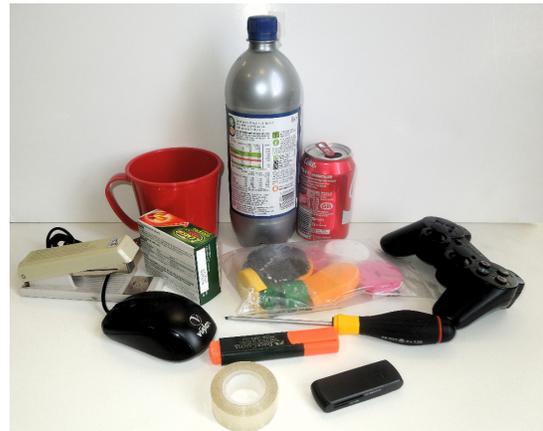}
      \caption{Object set used for the grasping experiments.}
      \label{fig: objects}
\end{figure}

The grasping experiments were carried out using 12 distinct household
objects (Fig. \ref{fig: objects}) and placed on one of three different
heights; changing in this way the approaching angle to execute the
grasp. The set of objects comprises of: stapler, pen marker, memory
stick, mug, bottle, tape roll, screwdriver, medicine pill box,
joystick, computer mouse, soda can and a bean bag.

\begin{figure}[thpb]
      \centering
      \includegraphics[width=0.4\textwidth]{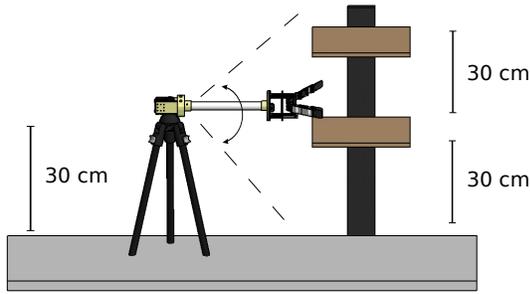}
      \caption{Grasping task setup. The hand is mounted on a tripod and objects are placed at three different heights (shelves).}
      \label{fig: setup}
   \end{figure}

The device is fixed to a tripod that allows to adjust and fix the
approaching angle/height of the grasp in a repeatable and controlled
fashion. The tripod's plate is set at 30 cm height and the three
``shelves'' are located at 0, 30 and 60 cm. Fig. \ref{fig: setup}
depicts the experiment setup.

With this setup, the experiment consisted in placing all objects (one
at a time) in 5 different orientations on each ``shelf'' level and
attempt grasps with each of the four hand configurations. A total of
720 combinations are obtained, performing 5 trials for each
combination to obtain a success rate, these parameters are listed in
Table \ref{tab: parameters}.  Each attempt was counted as successful
if the hand was able to keep grasp of the object when the arm raised
from the shelf level and the grip held the grasp for 15 seconds.  The
accompanying video shows the hand designs, setup and tests used and
Fig. \ref{fig: objects} shows the objects considered for the
experiment.

\begin{table}[h]
\caption{Experimental Parameters}
\begin{center}
\begin{tabular}{|c||c|c|c|}
\hline
Parameter & Min Value & Max Value & Samples\\
\hline
Orientation [degrees] & 0 & 359 & 5\\
\hline
Fingers & 2 & 4 & 4\\
\hline
Shelf height [cm] & 0 & 60 & 3\\
\hline
Object & - & - & 12\\
\hline
\end{tabular}
\end{center}
\label{tab: parameters}
\end{table}

\section{RESULTS AND DISCUSSION}
\label{section: results}

\begin{figure*}[thpb]
      \centering
      \includegraphics[width=\textwidth]{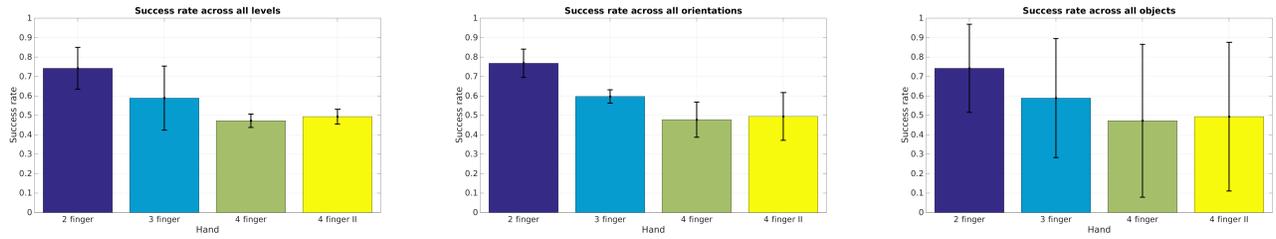}
      \caption{Success rate and variance averaged by shelf level (L), orientation (C) and object (R)}
      \label{fig: errors}
   \end{figure*}

Fig. \ref{fig: errors} shows the success rate of the four hand
configurations tested. Each plot includes error bars showing the
variance of the success rate according to shelf level, object and
object orientation respectively. Somewhat surprising is that the best
performance is achieved with the 2-fingered hand which achieves an
overall 75\% success rate.  On the other hand, the worst performance
is observed by the 4-fingered hands with a slight improvement in model
II of this configuration.

%   \begin{figure}[thpb]
%       \centering
%       \framebox{{\includegraphics[scale=.15]{per_hand}}}  
%       \caption{Overall success rate of every hand configuration.}
%       \label{fig: overall_hand}
%    \end{figure}

Looking at this data, we can see that for the average performance
across shelf levels the 2 and 3-fingered hands presents the highest
variance ($\sigma=.17$) which makes difficult to note a difference
between their performance. In contrast, both 4-fingered hands show low
success rates and little variance which confirms the poor behaviour
observed in most cases during the experiments. A first comparison can
be made using these values, for instance there appears to be a
notorious difference (25\% on average) between 2-fingers and both
4-fingered hands; not so between 2 and 3-fingered or between both
4-fingered hands.

Fig. \ref{fig: errors}(C) reflects a different scenario, where 2 and
3-fingered hands seemed less sensitive to changes in the orientation
in contrast to the 4-fingered configurations.  No important difference
is noted between the latter pair or between these and the 3-fingered
configuration. However, the 2-fingered hand shows a noticeable
performance difference.  From the plot can be seen that the all the
success rates are more than one standard deviation away from the
2-fingered score.

Regarding the average performance of the hand configuration across the
different objects there is no significant difference (big overlap in
error bars). Whereas the 2-fingered hand seems to be perform better on
average, it has a large error ($\sigma=.23$).

A multivariate logistic regression was fitted to assess the
significance of these results. Tab. \ref{tab: regression} shows the
outcome of this process. Here can be seen the relevance of the
parameters involved in the grasping task. The null hypothesis is that
the probability of a successful grasp is not associated with the value
of the parameters (hand configuration, object, shelf level,
orientation). 

Using the information from the table we can reject the
null hypothesis for all the parameters except for the type of object being
grasped; which as seen in Fig.\ref{fig: errors}(R) presents wider
variance. In other words, the grasping success related to shelf level and orientations changes is statistically significant to select a hand superior to the rest. In contrast, the information from object types shows no statistical proof to say that a hand works best for every type of object.

\begin{table}[ht]
\caption{Statistical significance of parameters}
\begin{center}
\begin{tabular}{lllll}
  \hline
 & Estimate & Std. Error & z value & Pr($>$$|$z$|$) \\ 
  \hline
(Intercept) & 22.8296 & 755.3303 & 0.03 & 0.9759 \\ 
  Hand 2 & -1.2845 & 0.3037 & -4.23 & $0.0000^{***}$ \\ 
  Hand 3 & -2.5779 & 0.3422 & -7.53 & $0.0000^{***}$ \\ 
  Hand 4 & -2.3535 & 0.3324 & -7.08 & $0.0000^{***}$ \\ 
  Marker & -19.3707 & 755.3302 & -0.03 & 0.9795 \\ 
  Box & -14.3254 & 755.3308 & -0.02 & 0.9849 \\ 
  Bottle & -0.0584 & 1068.2704 & -0.00 & 1.0000 \\ 
  Joystick & -19.4808 & 755.3302 & -0.03 & 0.9794 \\ 
  Screwdriver & -21.6513 & 755.3303 & -0.03 & 0.9771 \\ 
  Mouse & -19.4010 & 755.3302 & -0.03 & 0.9795 \\ 
  Beanbag & -19.8767 & 755.3302 & -0.03 & 0.9790 \\ 
  Memory stick & -20.2534 & 755.3302 & -0.03 & 0.9786 \\ 
  Mug & -14.3591 & 755.3308 & -0.02 & 0.9848 \\ 
  Stapler & -18.9864 & 755.3302 & -0.03 & 0.9799 \\ 
  Tape & -19.5005 & 755.3302 & -0.03 & 0.9794 \\ 
  Shelf & -0.9448 & 0.1466 & -6.45 & $0.0000^{***}$ \\ 
  Orientation & -0.0022 & 0.0011 & -1.94 & 0.0421 $\cdot$ \\ 
   \hline
   \hline
  %\hline 
 \textit{Significance:} &$\cdot \rightarrow 0.1$&$^{*} \rightarrow 0.05$& $^{**} \rightarrow 0.01$& $^{***} \rightarrow 0.001$
\end{tabular}
\end{center}
\label{tab: regression}
\end{table}

\subsection{Hand posture}

\begin{figure}[thpb]
      \centering
      \includegraphics[width=0.4\textwidth]{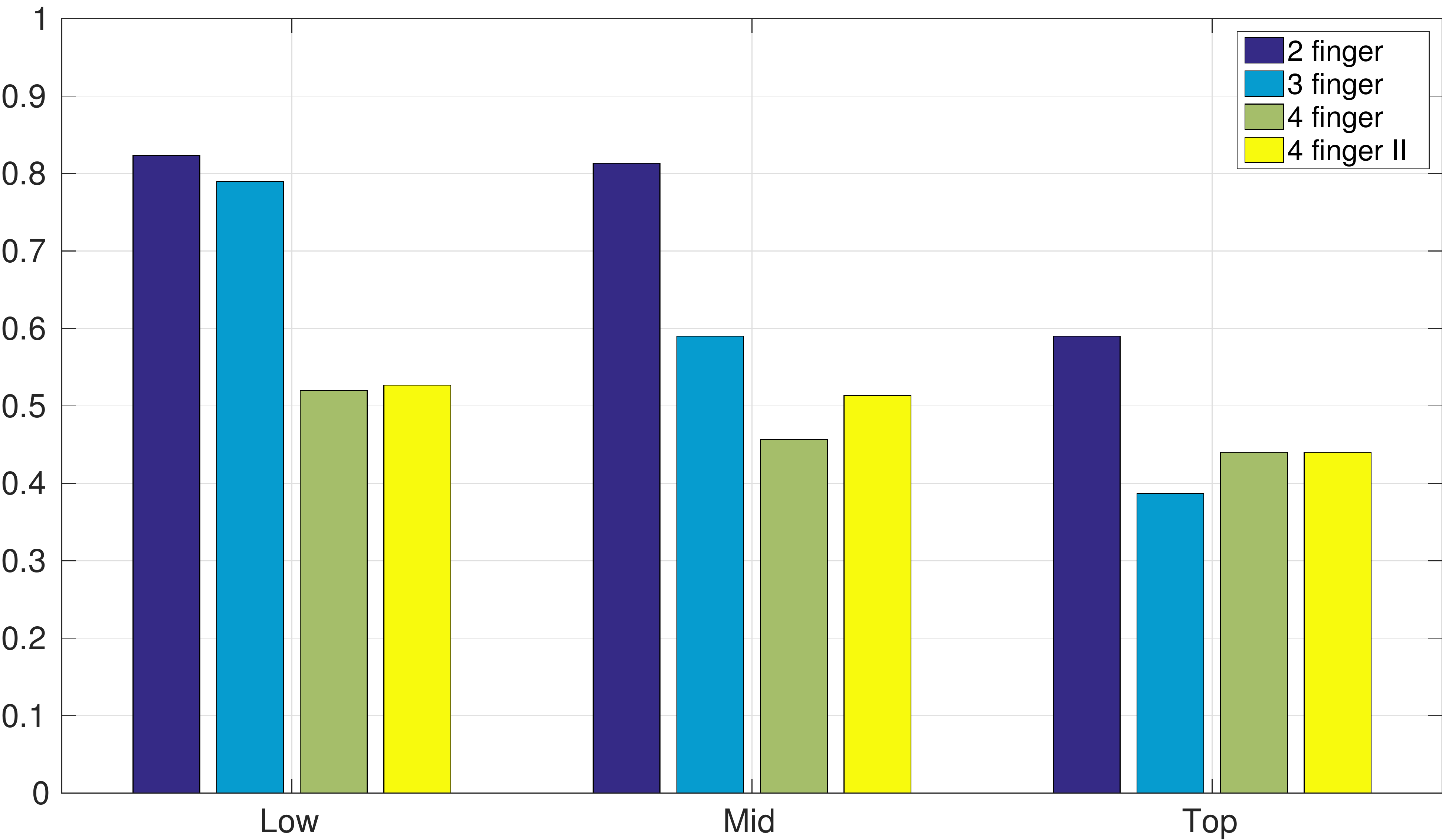}
      \caption{Average success of all hand configurations according to shelf level where the grasp is attempted.}
      \label{fig: per_level}
   \end{figure}

The hand posture is associated to the shelf level in which the grasp
is attempted, since a change in the height will be reflected in the
angle (posture) in which the hand approaches the
object. Fig. \ref{fig: per_level} shows the success rate of each hand
overall objects and orientations achieved at each shelf level in which
the grab was attempted.  For the low level shelf, similarities were
noticed between the performance of the 2 and 3-fingered hand
configurations, and between both 4-finger configurations. While the
former pair achieved around 80\% of successful grasp, the latter
remained at around only 50\%.

Concerning the middle level shelf, the 2-fingered and both of the
hands with 4 fingers showed roughly the same performance with respect
to the previous level. A more interesting results is observed in the
3-fingered hand, for which the success rate drops nearly 20\% with
relation to the previous shelf. Finally, the top level surface stood
as the most challenging level to grasp objects. At this level all the
hand configurations struggle to grab objects; while the two 4-fingered
hands kept achieving low scores, the 2 and 3-fingered hands
experienced a 20\% drop respect to the success rate obtained in the
previous level.

This plot shows that the worst outcome in all hand configurations is
obtained at the top level surface. The reason of this low success rate
was observed to be that many objects were out of reach for all
configurations. Due to the angle in which the fingers were located at
the start of the grasp, they fail to even touch the object. An example
of this situation for the 2-fingered hand is shown in Fig. \ref{fig:
  fail case}, which depicts a failed case when trying to grab a memory
stick at the top-level shelf. Due to the constrained wrist posture
small object generally were very few times grasped. 
While these conditions appear overly harsh, recall we are on purpose
not considering any planning but rather interested to evaluate
precisely performance in the absence of higher intelligence, or should
it be relevant, when planning has great uncertainty.

\subsection{Object pose}

The experiments carried out also showed the variation in the grasping
performance according to the objects pose. Fig. \ref{fig:
  per_orientation} details the success rate according to the object
orientation, averaged along the three shelf levels. Again the
2-fingered hand achieves the highest success rate across all
orientations. In an overall picture, all hands performed best when the
object has an orientation closer to 0 and 180 degrees (object's
largest axis was perpendicular to the palm).
% An example of this case is shown in Fig. \ref{fig: object_orientation_example}.

  \begin{figure}[thpb]
      \centering
      \includegraphics[width=0.4\textwidth]{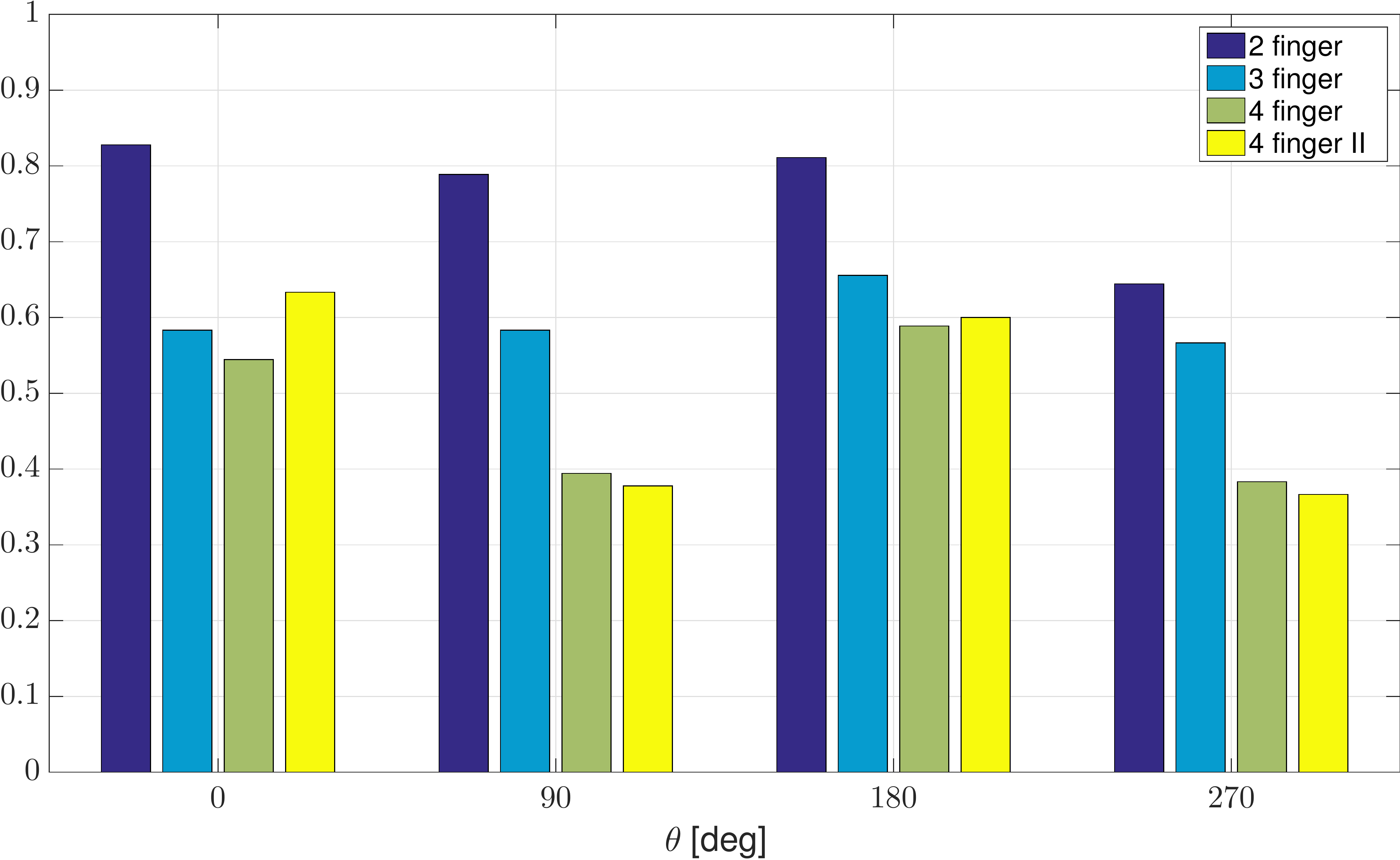}
      \caption{Average success per hand configuration according to the object orientation}
      \label{fig: per_orientation}
   \end{figure}
  
The worst performance across all hands was obtained when the objects
orientation was closer to 270 degrees. One might expect, as was the
case in 0 and 180 degrees, a similar performance between 270 and 90
degrees; while 3 out 4 hands showed this expected behaviour, for the
2-fingered hand that was not the case. The difference in that
performance (about 18\%) is associated to the irregularities in the
objects set shapes, strongly influenced by the performance at the
top-level surface. Another important thing to notice is that the
3-fingered configuration remains as the most consistent along all
orientations with a success rate around 60\%, its variance was
previously shown in Fig. \ref{fig: errors}.

\begin{figure*}[thpb]
      \centering
      \includegraphics[width=\textwidth]{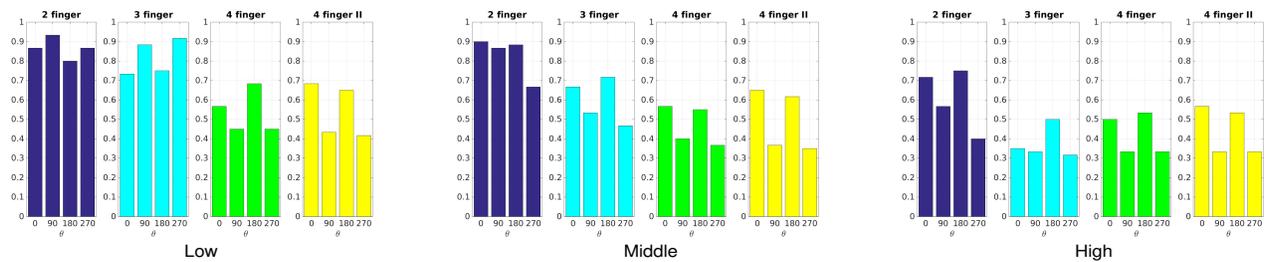}
      \caption{Average success of all hands for combinations of shelf level and object orientation. Low level shelf (L), Middle shelf(C) and High level shelf(R).}
      \label{fig: per_level_orientations}
   \end{figure*}

An wider view of the results is shown in Fig. \ref{fig:
  per_level_orientations}, which shows success rate for combinations
of shelf level and object orientation. From these graphs it can bee
seen that the 2-fingered hand's highest success rate was achieved at
the low-level shelf with objects oriented around 90 degrees; while the
worst performance resulted in a tie between the two 4-fingered hands
on the mid-level shelf with objects at 270 degrees. Whereas the low
performance does not results very surprising, the highest performance
does. In fact, the four cases with the highest performance are
observed when grasping objects at 90 degrees. This is related to the
fact that there are many objects with bladed shapes (flat and
elongated), which have a significant difference between the two axis
parallel to the shelf surface. When the longest axis is perpendicular
to the hand palm (90 and 270 degrees in our experiments) it offers
more contact possibilities when the grasp is attempted. Particularly,
it often was observed during the experiments that one finger made
contact with the object previously to the other(s), which made the
object spin or move. The grasp results less sensitive to this effect
when the fingers close along the object's longest axis.

% Fig. \ref{fig: matrices} shows the success score of each hand according to different combinations of the shelf level and object orientation. 
% \begin{figure*}[thpb]
%       \centering¹
%       \framebox{{\includegraphics[width=\textwidth]{matrices}}}  
%       \caption{Matrices}
%       \label{fig: matrices}
%    \end{figure*}

\subsection{Performance by object}

\begin{figure}[thpb]
      \centering
      \includegraphics[width=0.4\textwidth]{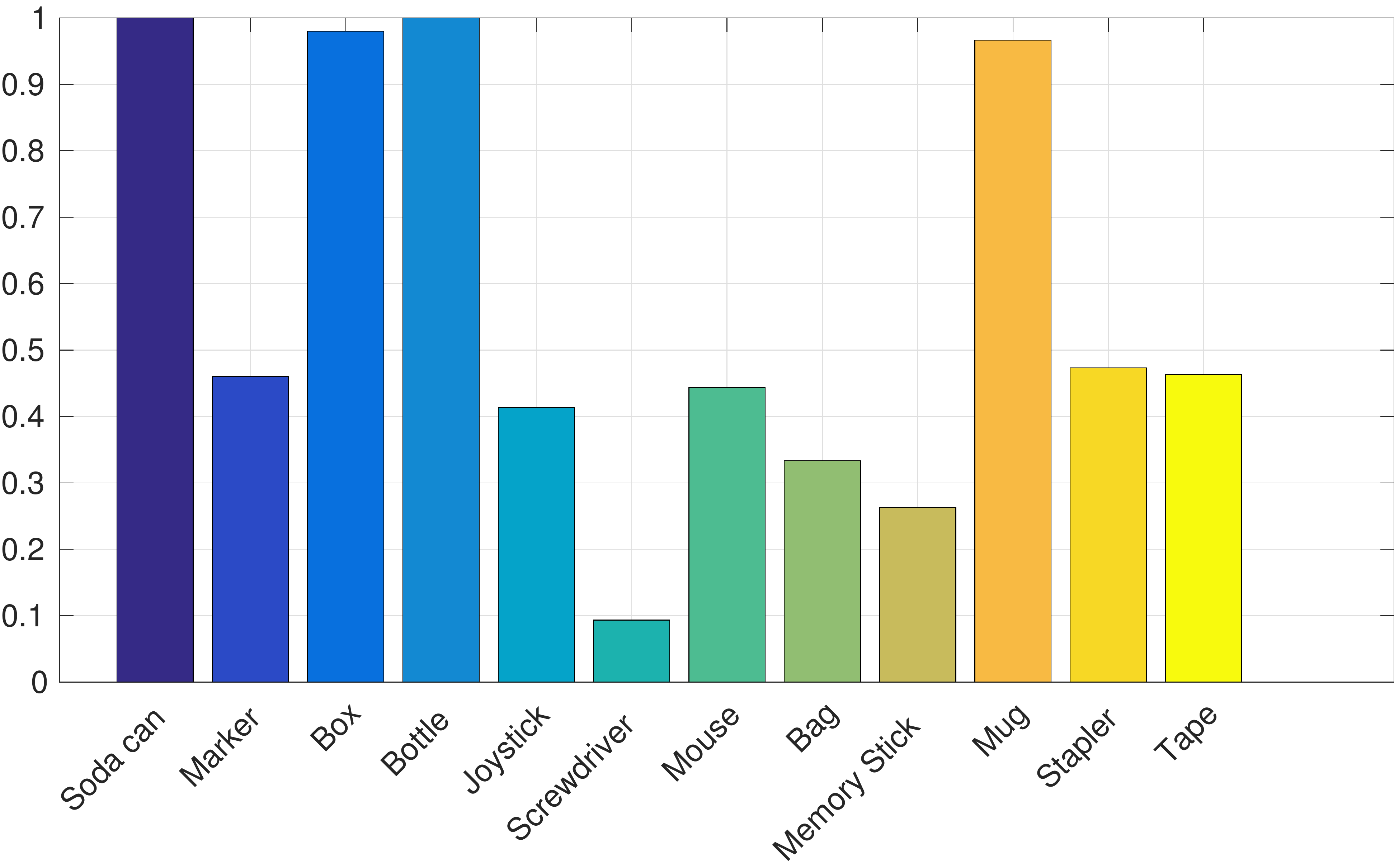}
      \caption{Average success for each object across all trials }
      \label{fig: per_object}
   \end{figure}

From previous subsection has been highlighted the large variance in
the grasping performance due to the diversity of object's
shapes. Fig. \ref{fig: per_object} shows the success rate achieved by
each object considered for the experiments. This performance was
averaged across all hands; however, it gives insight into what happens
when attempting to grasp the various objects.

It is curiously noted that objects usually employed in the robotic
grasping literature for demonstrations e.g. soda can, bottle, mug and
box, achieved in our experiments the highest success rates. This
subset of objects is often considered when hand designs are tested and
grasping algorithms proposed. As can be seen in Fig. \ref{fig:
  per_object_per_hand}, these objects did not represent a challenge
for any of the hand configurations put to test, which often achieve a
success rate above 80\%. On the other hand, objects such as the
screwdriver or the memory stick represent challenging situation in the
majority of cases. We have included objects of different classes,
sizes, geometries and weights which we believe is a better
approximation of what can occur in a realistic environment. It is
worth noticing that the score of the beanbag is within one standard
deviation ($\sigma=.2$) from the average performance of the hand, more
precisely 8\% under the average which is not as low as expected,
taking into account that this object is non-rigid or deformable.

\begin{figure}[thpb]
      \centering
      \includegraphics[width=0.4\textwidth]{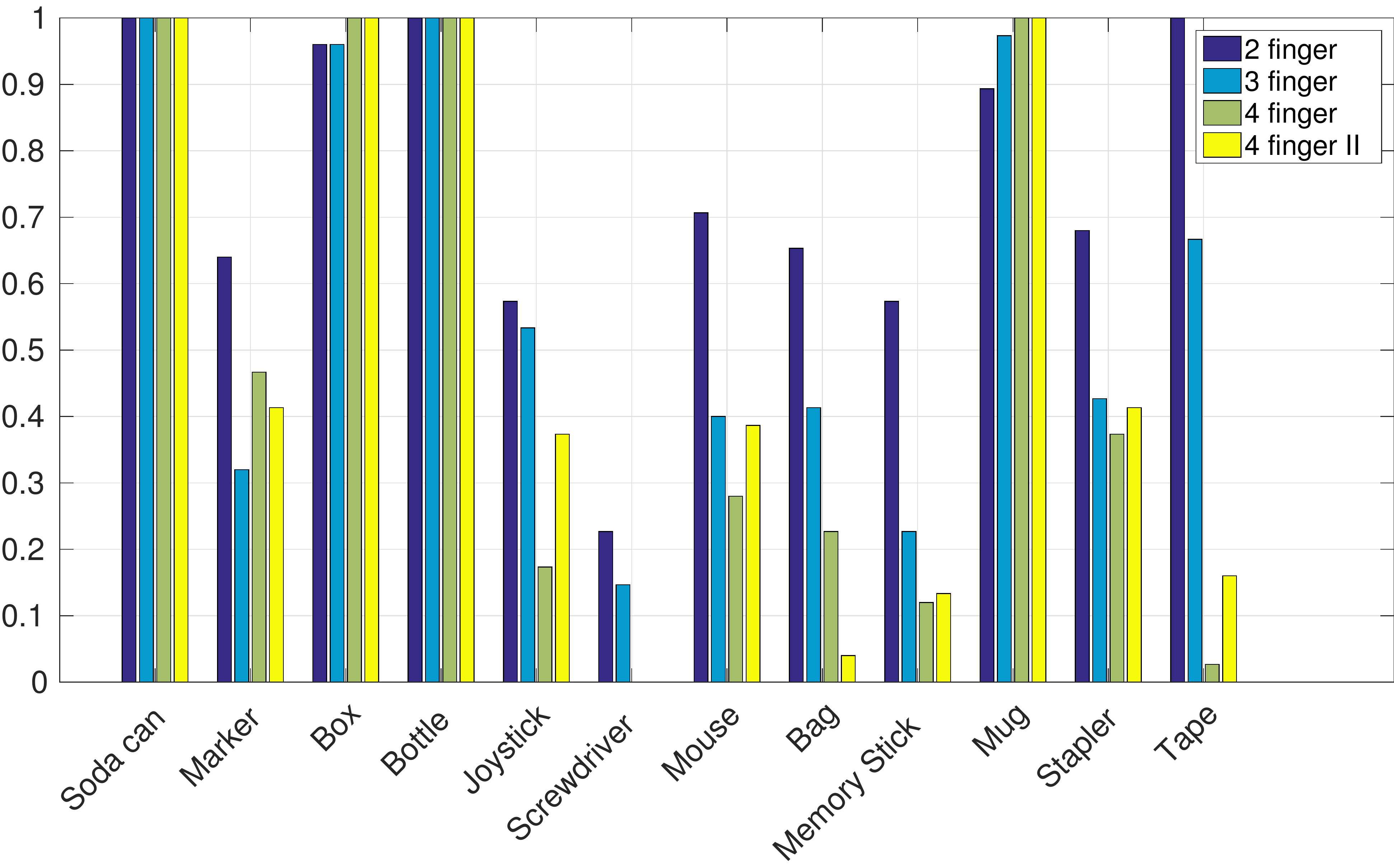}
      \caption{Average success achieved by each hand for every object in the experiment. }
      \label{fig: per_object_per_hand}
   \end{figure}

As a final analysis we compared each hand using a statistical model in
order to see the significance of the grasping parameters per
hand. This approach is similar to the one applied to the complete
grasping data in the first part of this section. Using logistic
regression we can assess to what extent (statistically) the change in
shelf level, type of object and orientation influence a successful
grasp. Table \ref{tab: regression 2} shows the \textit{p}-values for
the statistical significance of the parameters for each hand.

  \begin{figure*}[thpb]
      \centering
      \includegraphics[width=\textwidth]{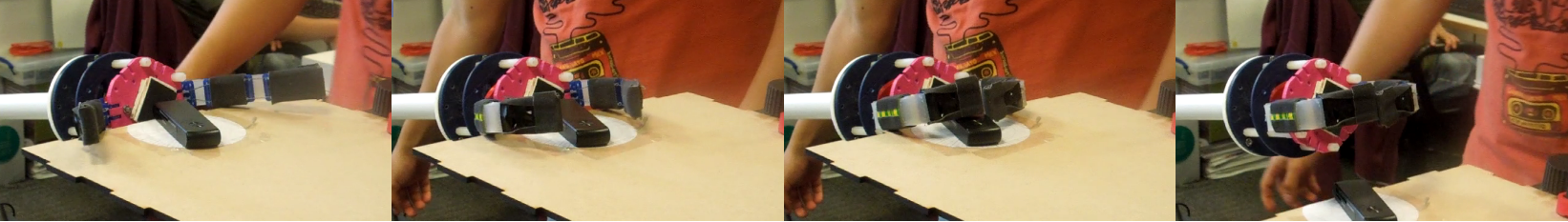}
      \caption{Example of a failure case on the top level shelf. The object (memory stick) is out of the reachable workspace of the hand. This usually happened to small objects on the
      the highest shelf.}
      \label{fig: fail case}
  \end{figure*}
  
\begin{table}[ht]
\caption{Statistical significance of parameters by hand}
  \begin{center}
\begin{tabular}{lllll}
  \hline
 & Pr($>$$|$z$|$) & Pr($>$$|$z$|$)  & Pr($>$$|$z$|$) & Pr($>$$|$z$|$) \\ 
  \hline
(Intercept)  & 0.9927 & 0.9921 & 0.9962 & 0.9960\\ 
  Marker  & 0.9939 & 0.9930 & 0.9963 & 0.9962\\ 
  Box  & 1.0000 & 0.9945 & 1.0000 & 1.0000\\ 
  Bottle  & 1.0000 & 1.0000&  1.0000 & 1.0000\\ 
  Joystick  & 0.9939 & 0.9935 & 0.9960 & 0.9962\\ 
  Screwdriver  & 0.9933 & 0.9926 & 0.9949 & 0.9947\\ 
  Mouse  & 0.9941 & 0.9931 & 0.9961 & 0.9962\\ 
  Beanbag  & 0.9941 & 0.9932 & 0.9961 & 0.9947\\ 
  Memory stick  & 0.9940 & 0.9926 & 0.9960 & 0.9959\\ 
  Mug & 0.9947 & 1.0000 & 1.0000 & 1.0000\\ 
  Stapler & 0.9940 & 0.9932 & 0.9963 & 0.9963\\ 
  Tape & 1.0000 & 0.9937 & 0.9949 & 0.9947\\ 
  Shelf   & $0.0000^{***}$ & $0.0000^{***}$ & 0.3487  & 0.1196\\ 
  Orientation   & 0.1175 & 0.6472 & 0.2639 & $0.0476^*$\\
   \hline
   \hline
 \textit{Significance:} &$\cdot \rightarrow 0.1$&$^{*} \rightarrow 0.05$& $^{**} \rightarrow 0.01$& $^{***} \rightarrow 0.001$
\end{tabular}
\end{center}
\label{tab: regression 2}
\end{table}

From Table \ref{tab: regression 2} can be observed that for the 2 and
3-fingered hands the change in shelf levels is highly significant
whereas for the 4-fingered one there is no significant proof of how
the parameters affect the performance. In contrast, the second model
of the 4-fingered hand seemed more sensitive to changes in the objects
orientation. The difference among hands is more clear when these data
is compared with the overall performance. For instance, the 4-fingered
configurations show low (or none) statistical dependence on the task
parameters but when taking into consideration their low performance
conclusions can be made about the effectiveness of such
designs. Similarly, statistical analysis shows that the 2 and
3-fingered success rates are actually related to the task parameters
in similar ways (\textit{p}-values); however when the overall
performance is taken into account the 2-fingered design becomes
superior most of the time.

\section{CONCLUSIONS}
\label{section: conclusions}

In the search for a multi-purpose end-effector we have evaluated
variations of a state-of-the-art underactuated compliant hand. With
the introduction of a few adaptations, it has been put to test using
multiple finger configurations and an object set containing various
types of objects; in what we believe is a common scenario for robotic
grasping. While this type of hands have had success in performing
grasp with ``common'' objects in the past, we have found that a
general-purpose hand, remains elusive. However, we believe that the
results obtained here raise interesting questions about hand design
and performance evaluation. For instance, is a high number of fingers
relevant with the introduction of underactuated compliant designs? Our
results obtained from 3600 object-hand interactions point to design
simplicity of only two fingers even in the absence of any planning
strategy.

%\section*{ACKNOWLEDGMENT}


%CONACYT?

%%%%%%%%%%%%%%%%%%%%%%%%%%%%%%%%%%%%%%%%%%%%%%%%%%%%%%%%%%%%%%%%%%%%%%%%%%%%%%%%

\bibliography{IEEEabrv,refe}
\bibliographystyle{IEEEtran}

\end{document}